\documentclass{article}
\usepackage{iclr2027_conference,times}
\usepackage{iftex}

\usepackage{amsmath,amsfonts,bm}

\def\eqref#1{equation~\ref{#1}}

\def\1{\bm{1}}

\DeclareMathAlphabet{\mathsfit}{\encodingdefault}{\sfdefault}{m}{sl}
\SetMathAlphabet{\mathsfit}{bold}{\encodingdefault}{\sfdefault}{bx}{n}

\usepackage{graphicx}
\usepackage{booktabs}
\usepackage{makecell}
\usepackage{xcolor}
\usepackage{hyperref}
\usepackage{url}
\usepackage{float}

\title{Robotic Spatiotemporal Editing: A Unified Interface for Controllable Action Generation via Masked Action Reconstruction}
\iclrfinalcopy
\author{\parbox{\dimexpr\textwidth-2\tabcolsep\relax}{\centering\normalfont\linespread{1.1}\selectfont
\vspace{12pt}
\mbox{\textbf{Yanbang Li}\textsuperscript{1,2,*}}, \mbox{\textbf{Bu He}\textsuperscript{1,*}}, \mbox{\textbf{Yi Tian}\textsuperscript{1}}, \mbox{\textbf{Yu-Chiao Huang}\textsuperscript{1}}, \mbox{\textbf{Chenyu Zhang}\textsuperscript{1,3}}, \mbox{\textbf{Ruoqu Chen}\textsuperscript{1,3}}, \mbox{\textbf{Yuanzhe Liu}\textsuperscript{4}}, \mbox{\textbf{Yuxiang Gao}\textsuperscript{2}}, \mbox{\textbf{Mengdi Xu}\textsuperscript{1,\textdagger}}\\[5pt]
\mbox{\textsuperscript{1}Tsinghua University}\quad \mbox{\textsuperscript{2}CocoMatrix}\quad \mbox{\textsuperscript{3}Shanghai Qizhi Institute}\quad \mbox{\textsuperscript{4}UIUC}\\[3pt]
\textsuperscript{*}Equal contribution\quad \textsuperscript{\textdagger}Corresponding author\\[5pt]
{\hypersetup{pdfborder={0 0 0}}\href{https://robotic-spatiotemporal-editing.github.io/rose.github.io/}{\textcolor{blue}{\nolinkurl{https://robotic-spatiotemporal-editing.github.io/rose.github.io/}}}}
}}
\hypersetup{pdfauthor={}}

\begin{document}
\maketitle

\lhead{}
\vspace{-5pt}
\begin{abstract}
Generative policies can synthesize robot behaviors, yet controlling and editing them through diverse spatiotemporal references remains challenging.
Real-world manipulation requires robots to satisfy user-specified constraints on where, when, and how actions are executed. 
Our key insight is that spatiotemporal control references, such as keypoints, flows, orientations, and keyposes, can be viewed as partial specifications of the same temporal Cartesian action space.
We introduce the \textbf{MAP (Masked Action Space)} interface to unify these action references and formulate controllable manipulation as an action inpainting problem.
Our \textbf{Inpainting Policy (IP)} reconstructs missing action dimensions of the end-effector, combining progress-guided alignment with multi-mask training, supporting unseen spatiotemporal constrains.
We further adopt a training-free \textbf{RePaint} variant that iteratively inpaints the action during denoising to reduce control error.
On ManiSkill and LIBERO-10, our policy achieves average task success rates of $92.67\%$ and $64.67\%$, respectively.
Real-world experiments demonstrate spatiotemporal control with average positional and rotational errors of \(44.3\,\mathrm{mm}\) and \(7.8^\circ\), respectively.
\vspace{-2pt}
\end{abstract}

\section{INTRODUCTION}
\label{sec:introduction}
\vspace{-8pt}
\begin{figure}[H]
    \centering
    \includegraphics[width=\linewidth]{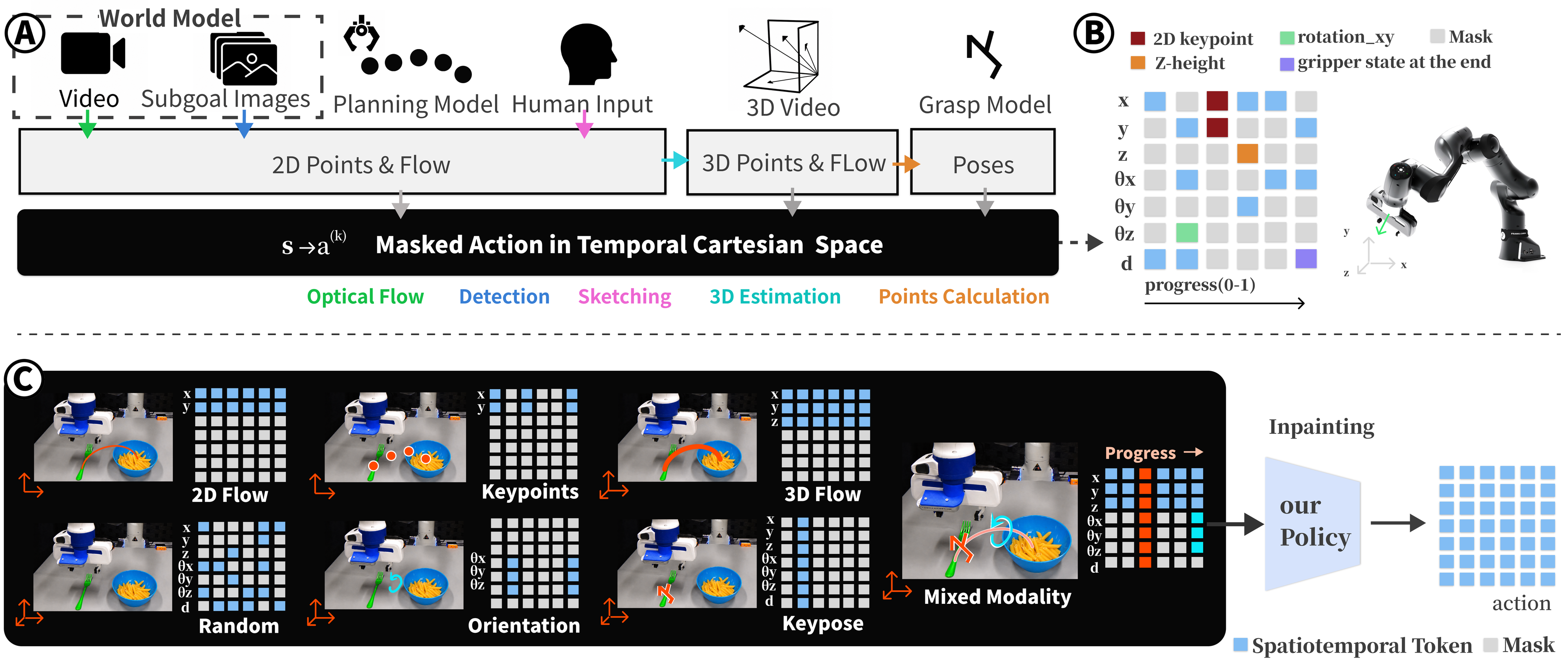}
    \vspace{-10pt}
    \caption{\textbf{Unified spatiotemporal control via action inpainting.}
    \textbf{Top:} To enable unified spatiotemporal control from heterogeneous model-generated and human-provided references, we introduce \textbf{Masked Action Space (MAP)}, which represents these references as partial specifications of temporal Cartesian actions. Progress values specify when constraints apply, instead of timestep.
    \textbf{Bottom:} Our \textbf{Inpainting Policy (IP)} predicts these masks through action inpainting, supporting potentially
    unlimited spatiotemporal reference types and unseen reference combinations.}
    \label{fig:masked_cartesian_action_space}
\end{figure}
\vspace{-4pt}
Generative models have recently demonstrated remarkable progress in robotic manipulation, enabling policies to synthesize complex robot behaviors from visual observations and task descriptions~\citep{chi2025diffusion,black2026pi0visionlanguageactionflowmodel, brohan2023rt1roboticstransformerrealworld}.
However, generating plausible behaviors alone is insufficient for deploying robots in the physical world.
Real-world tasks require robots not only to decide what to do but also to satisfy constraints on \textbf{where, when, and how} to execute an action ~\citep{rempe2026kimodo}.
Moreover, humans and intelligent systems often need to refine, modify, and compose robot behaviors by adding spatiotemporal constraints.
For example, when tidying a table, a user may use a hand gesture or a sketch instruction~\citep{li2025roboticvisualinstruction} to specify where an object should be placed, rather than merely asking the robot to tidy the table.

How can we edit robot behaviors in space and time? In image and video generation, the shift from generating content from user instructions to enabling fine-grained, controllable editing has reshaped image and video generation industry. Reference-based interfaces, such as ControlNet~\citep{zhang2023addingconditionalcontroltexttoimage}, enable users to steer synthesis through interpretable signals like edges, depth, and poses. They demonstrate that a well-designed interface can bridge human's spatial intent and generative outcomes.

Meanwhile, robotics is gaining access to diverse spatiotemporal references from world models, 3D/4D videos, advanced agents, grasp models, and human demonstrations~\citep{jang2025dreamgenunlockinggeneralizationrobot,gao2026dreamdojogeneralistrobotworld,chen2026dexxlearningvisualtactiledexterous,murali2025graspgendiffusionbasedframework6dof,fan2025longvlaunleashinglonghorizoncapability}, including keypoints~\citep{huang2024rekep}, flow~\citep{xu2024flowcrossdomainmanipulationinterface}, keyposes~\citep{yu2024bikckeyposeconditionedconsistencypolicy}, and grasp proposals~\citep{fang2023anygrasp,hao2025disco} (see Figure~\ref{fig:masked_cartesian_action_space}(A)).
These representations are physically grounded and low-dimensional, but they only provide partial information on future robot motion rather than complete executable actions.
Existing approaches typically address this gap with signal-specific policies or conditioning mechanisms.
This motivates the need for \textbf{a unified action interface} that can transform diverse spatiotemporal references into controllable robot behaviors.

Our key insight is that heterogeneous spatiotemporal control are not different control problems, but partial action tokens of the same underlying Temporal Cartesian Action Space.
We design \textbf{Masked Action Space (MAP)} as a shared interface that represents these references through partially specified action tokens and a binary mask (see Figure~\ref{fig:masked_cartesian_action_space}(B,C)).
We formulate controllable action generation as \textbf{a masked action inpainting problem}, and we propose \textbf{Inpainting Policy (IP)} to reconstruct masked components in MAP and turn them into a complete executable trajectory (see Section~\ref{sec:inpainting_policy}).

To enable this formulation, we design three components.
\textbf{(1)} A \textbf{progress-guided sliding window} aligns control references with the current task stage using estimated progress value~\citep{chen2026sarmstageawarerewardmodeling} rather than raw execution timesteps (Section~\ref{sec:progress_guided_window}).
\textbf{(2)} A \textbf{two-stage joint-training strategy} uses random masks in the first stage and structured masks in the second stage to learn robust control over diverse and unseen control references (Section~\ref{sec:multi_modal_training}).
\textbf{(3)} As a \textbf{training-free} alternative, we also adopt a \textbf{RePaint-style inpainting strategy} that can be applied during the inference-time denoising process of an off-the-shelf policy.
We conduct experiments on 15 tasks across the ManiSkill, LIBERO-10, and real-world environments, covering six standard control references and randomly structured references.
In sum, our method demonstrates several capabilities:

\begin{itemize}
\item \textbf{Versatile controllability.}
Inpainting Policy supports diverse spatiotemporal references, including keypoints, flows, orientations, poses, and combinations of these reference types.
In real-world experiments, it attains mean position and rotation errors of \(44.3\,\mathrm{mm}\) and \(7.8^\circ\), respectively.
Meanwhile, our method improves the success rate on LIBERO-10 by 4.67 percentage points compared with the baseline.
This proves that our policy enables arbitrary spatiotemporal controllability without compromising task completion.

\item \textbf{Interpretability.}
MAP (Masked Action Space) provides a human-readable and editable interface that bridges multi-dimensional spatiotemporal information from the physical world with robotic manipulation actions via the masked temporal Cartesian space.
It explicitly indicates which timesteps and action dimensions are constrained.

\item \textbf{Practicality.}
Our method can be directly applied to diffusion- or flow matching-based policies in simulated or real-world environments.
Built on a lightweight \(49\mathrm{M}\)-parameter DiT model, it achieves a \(92.67\%\) success rate and a control error of \(0.023\) (Table~\ref{tab:multitask_single_task_500}) in ManiSkill experiments.
In real-world experiments, a \(118\mathrm{M}\)-parameter diffusion model achieves a success rate above \(90\%\).
\end{itemize}

\section{RELATED WORK}
\label{sec:related_work}

\subsection{HETEROGENEOUS SPATIOTEMPORAL SIGNALS FOR ROBOTICS CONTROL}

In learning-centric robotics research, actions are increasingly encoded as low-dimensional spatio-temporal signals because they are interpretable and spatially grounded.
Among such signals, point-based constraints~\citep{huang2024rekep,liu2024moka} constitute spatially sparse guidance and commonly represent waypoints or affordance-based grasp points~\citep{fang2025kalm}. Moreover, vision--language models (VLMs), such as Gemini Robotics-ER~\citep{team2025gemini} and Qwen~\citep{qwen2025qwen25technicalreport}, have been integrated into several VLA frameworks~\citep{xu2026a0affordanceawarehierarchicalmodel,ji2025robobrain} to output waypoint sequences that guide action planning.
Another line of work employs keyposes as structural anchors of motion, first forecasting intermediate keyposes and then reconstructing full trajectories via downstream action models~\citep{fang2025kalm,yu2024bikckeyposeconditionedconsistencypolicy,yang2025gripper,fang2023anygrasp}. Beyond spatially sparse signals, trajectories provide spatio-temporally dense guidance and are widely used to condition downstream action generation~\citep{xu2024flowcrossdomainmanipulationinterface,wen2024anypointtrajectorymodelingpolicy,gu2023rt,liu2025gae}.
In parallel, video generation and dynamics models~\citep{assran2025v,liu2024sora} have been repurposed for planning by predicting future visual scenes~\citep{li2025unifiedvideoactionmodel}.
Despite their effectiveness, current designs typically rely on modality-specific action policies, hindering integration and generalization across spatio-temporal conditions for action generation. Our work targets this gap by seeking a unified robot learning method that standardizes the conditioning interface across heterogeneous spatio-temporal signals.

\subsection{INPAINTING FOR ROBOT CONTROL}

Inpainting methods are used widely in visual spaces. They using image- or video-level
masking and inpainting to augment observations, bridge embodiment gaps,
or represent actions: ROSIE~\citep{yu2023scalingrobotlearningsemantically}, Mirage~\citep{chen2024mirage},
Cloak~\citep{piseno2026cloak}, and Masked Visual Actions
~\citep{alzayer2026masked}.
Recent works on controllable human and humanoid motion generation include
Kimodo~\citep{rempe2026kimodo}, which imputes pose features,
Scheduled Inpainting~\citep{agrawal2026interactive}, which
balances preservation and editing, and MotionBricks
~\citep{wang2026motionbricks}, which performs latent in-betweening.
In manipulation action modeling, inpainting has primarily been used not for
purposes on controllable action generation. RTC~\citep{black2506real} and DiscreteRTC~\citep{wang2026discretertc}
generate action chunks to accelerate asynchronous inference. In addition, HDP
~\citep{ma2024hierarchical} and DISCO~\citep{hao2025disco} use
inpainting to impose trajectory-endpoint and semantic-keyframe
constraints, respectively. To the best of our knowledge, \textbf{controllable action generation in space and time via action inpainting} remains largely unexplored for robotic manipulation.

\section{METHOD}
\label{sec:method}
Our method first introduces a unified controllable interface, the \textbf{MAP (Masked Action Space)}, for manipulation action generation, which converts heterogeneous spatiotemporal signals into specified action tokens and associated binary masks, thereby representing different control references in a common masked Cartesian action space (Section~\ref{sec:problem_formulation}).
Based on this representation, we develop an \textbf{Inpainting Policy} that reconstructs the masked action tokens and generates executable action chunks (Section~\ref{sec:inpainting_policy}).
Our framework consists of three components.
\textbf{(1)} A progress-guided sliding window aligns MAP with the current task progress and provides a temporally anchored MAP condition (Section~\ref{sec:progress_guided_window}).
\textbf{(2)} Multi-modal joint training over heterogeneous mask families enables structured control and zero-shot generalization to unseen spatiotemporal compositions (Section~\ref{sec:multi_modal_training}).
\textbf{(3)} We adopt RePaint~\citep{lugmayr2022repaintinpaintingusingdenoising} as a training-free inference strategy, which repeatedly reinserts the specified action tokens during the denoising process of the diffusion-based policy, thereby improving adherence to the provided constraints (Section~\ref{sec:inpainting_repaint}).

\subsection{MASKED ACTION SPACE}
\label{sec:problem_formulation}

We formulate reference-guided robotic manipulation as reconstructing complete actions from partial specifications in a temporal Cartesian action space.

We use normalized task progress \(p'\in[0,1]\) as the temporal coordinate, with reference alignment detailed in Section~\ref{sec:progress_guided_window}.
Let \(A^{\mathrm{ref}}\in\mathbb{R}^{T_{\mathrm{ref}}\times7}\) denote the reference action-value array at discrete progress locations \(p_p'\), where \(p=0,\ldots,T_{\mathrm{ref}}-1\).
At each location, the action vector is defined as \(a_p=(x_p,y_p,z_p,\theta_{x,p},\theta_{y,p},\theta_{z,p},d_p)\), where the first six dimensions describe the gripper pose and \(d_p\) denotes the gripper state.
Each scalar action entry is a spatiotemporal action token.
The spatiotemporal references are represented by a binary mask \(M\in\{0,1\}^{T_{\mathrm{ref}}\times7}\) and the corresponding specified action tokens \(M\odot A^{\mathrm{ref}}\), where \(M_{p,d}=1\) indicates a supplied value and \(M_{p,d}=0\) indicates an unspecified entry.

\begin{figure}[t]
\centering
\includegraphics[width=\linewidth]{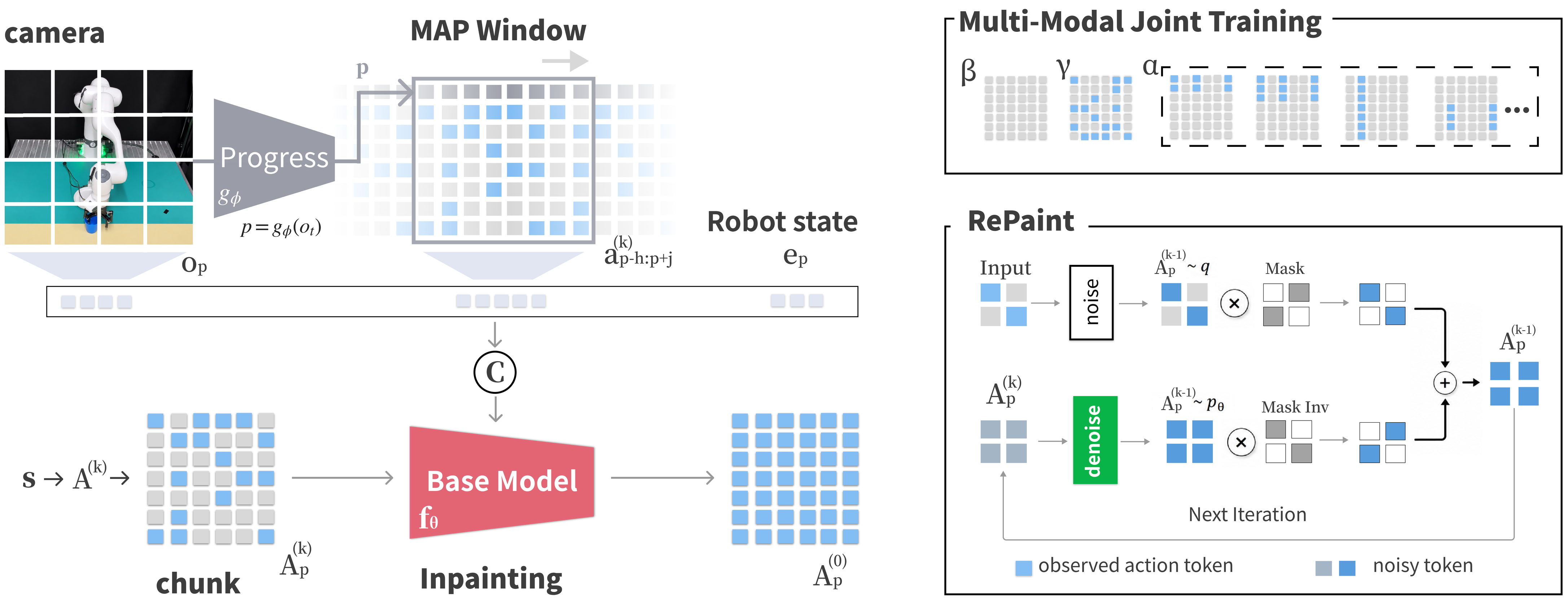}
\caption{\textbf{Inpainting Policy}.
Given camera observations and proprioceptive information, a progress network anchors a local MAP window and masked action chunk in the temporal Cartesian action space.
The window is encoded into a feature vector to condition the base model, which predicts missing action tokens from the masked action chunk.
Multi-modal joint training enables the policy to generalize to unseen spatiotemporal modalities.}
\label{fig:inpainting_policy}
\end{figure}

\paragraph{Mapping Control References to Masks.}
The mapping \(s\to A^{(k)}\) converts control references \(s\) into MAP constraints on the action chunk (Figure~\ref{fig:inpainting_policy}).
As illustrated in Figure~\ref{fig:masked_cartesian_action_space}(C), different distributions of \(M\) instantiate different control signals.
These references can be continuous or discrete along the temporal axis.
Structured mask types include 2D/3D keypoints, 2D/3D flow, keyposes, orientations, gripper states, and mixed-modality conditions formed by randomly combining these references.
Here, 2D references specify Cartesian \(x,y\) coordinates and flow references denote position trajectories; image-space references must first be mapped into the action coordinate system.
For a reference at progress \(p'\), let \(p^\star=\arg\min_p |p_p'-p'|\) denote its discrete reference index.
For example, a 3D keypoint activates \(M_{p^\star,\{1,2,3\}}=1\), while an orientation reference activates \(M_{p^\star,\{4,5,6\}}=1\).
MAP also supports unstructured mask types, as shown in Figure~\ref{fig:masked_cartesian_action_space}(B), such as specifying the gripper state at a particular manipulation stage, constraining the end-effector state along the \(z\)-axis, or specifying a rotation angle on a particular plane.

\subsection{PROGRESS-GUIDED SLIDING WINDOW OF MAP}
\label{sec:progress_guided_window}

We use normalized task progress \(p'\in[0,1]\) to align references with policy execution because execution speed can vary across rollouts, making raw timesteps unreliable for identifying corresponding task stages (see the related ablation study in Supplementary Section~1.5).
Given the current observation \(o_t=(I_t,e_t)\), where \(I_t\) denotes camera observations and \(e_t\) denotes the robot state, a progress network predicts \(p_t'=g_{\phi}(o_t)\in[0,1]\).
The corresponding reference index is \(p_t=\arg\min_p |p_p'-p_t'|\).

The progress estimate selects a local reference window \(\mathcal{W}_t=[a_{p_t-h},\ldots,a_{p_t},\ldots,a_{p_t+j}]\), where \(h\) and \(j\) specify the past and future context lengths.
The encoder receives only the specified entries and their visibility masks; boundary handling and window-length conventions are detailed in Supplementary Section~3.1.
The window associates reference constraints with the current task stage and provides nearby reference context for action generation.
Its temporal coverage is controlled by the backward and forward window lengths, whose effect is evaluated in Supplementary Section~1.2.

\subsection{INPAINTING POLICY}
\label{sec:inpainting_policy}

As illustrated in Figure~\ref{fig:inpainting_policy}, IP generates complete Cartesian action chunks conditioned on visual observations, robot state, and the local MAP window selected in Section~\ref{sec:progress_guided_window}.
A 2D convolutional encoder maps the local MAP window to a 128-dimensional feature vector, which is concatenated with visual and proprioceptive features to condition the denoising network (see Supplementary Sections~3.6 and~1.1 for architecture details and encoder comparisons).
Let \(A^{(0)}\in\mathbb{R}^{H\times7}\) denote a clean action chunk and \(A^{(k)}=\sqrt{\bar{\alpha}_k}A^{(0)}+\sqrt{1-\bar{\alpha}_k}\epsilon\) its noisy version at diffusion step \(k\), where \(\bar{\alpha}_k\) is the cumulative signal-retention coefficient and \(\epsilon\sim\mathcal{N}(0,I)\).
The denoising network predicts \(\widehat{\epsilon}=f_{\theta}(A^{(k)},k,C_t)\), where \(C_t\) contains the observation features and the encoding of \(\mathcal{W}_t\).
Let \(M_t^{\mathrm{chunk}}\in\{0,1\}^{H\times7}\) denote the reference visibility mask aligned with the target action chunk.
The training objective is
\[
\mathcal{L}_{\mathrm{MAP}}
=
\mathbb{E}
\left[ w_m \left\| M_t^{\mathrm{chunk}} \odot (\epsilon-\widehat{\epsilon}) \right\|_F^2+ w_u \left\| (1-M_t^{\mathrm{chunk}}) \odot (\epsilon-\widehat{\epsilon}) \right\|_F^2
\right],
\]
where \(w_m\) and \(w_u\) control the relative importance of known and unknown regions, and the expectation is over training examples, sampled masks, diffusion steps, and Gaussian noise.
In practice, we use a weighted reconstruction objective to encourage consistency with specified references while learning to reconstruct unspecified actions.

\subsection{MULTIMODAL JOINT TRAINING}
\label{sec:multi_modal_training}

To enable one model to handle diverse spatiotemporal conditions, we construct a multi-modal masked action dataset from demonstrations.
As illustrated in Figure~\ref{fig:inpainting_policy}, we train the Inpainting Policy with three complementary mask patterns: \(\alpha\), \(\beta\), and \(\gamma\), each targeting a distinct aspect of masked action reconstruction.

The \(\alpha\)-pattern consists of structured spatiotemporal control signals, including 2D keypoints/flow, 3D keypoints/flow, keyposes, orientation, and gripper-state conditions.
These masks represent common controllable action-generation interfaces, where partial Cartesian action tokens are specified as subgoals or constraints.
Joint training over \(\alpha\) enables the policy to follow diverse structured conditions within a unified action space.

The \(\beta\)-pattern corresponds to the no-known-context case, where no action reference is provided.
In this setting, the model reduces to plain action generation by the diffusion policy.
Including \(\beta\) encourages robustness when test-time conditions are sparse, such as low-frequency control signals.

The \(\gamma\)-pattern introduces random token-level masks.
Since structured masks are special cases of random masking, \(\gamma\) promotes token-level generalization across MAP, especially under OOD patterns.
However, experiments show that random-mask training may underperform structured joint training when training data is limited.

We sample a mask family \(c\sim\rho(c)\), where \(c\in\{\alpha,\beta,\gamma\}\), and then draw a mask \(M\sim p(M\mid c)\).
The distribution \(\rho(c)\) controls the relative frequency of structured, no-reference, and random-mask examples during policy training.
Mask families and compositions are detailed in Supplementary Sections~3.4--3.5.
For LIBERO-10, we first train with random masks and then continue with a mixture of structured masks, as detailed in Supplementary Sections~2.2 and~1.4.

We find that, after joint training, the model can follow compositions of different spatiotemporal constraints and generaloize to an OOD mask pattern, formed by combining several standard patterns but never observed during training (Figure~\ref{fig:mask-type-breakdown}).

\subsection{INPAINTING VIA REPAINT}
\label{sec:inpainting_repaint}

We adopt a RePaint-style inference strategy~\citep{lugmayr2022repaintinpaintingusingdenoising} that repeatedly reinserts specified reference values during denoising without additional policy training.
Given clean reference values \(R_t\in\mathbb{R}^{H\times7}\) aligned with the target action chunk, we sample their noisy version through forward diffusion.
The binary mask \(M_t^{\mathrm{chunk}}\) specifies the observed action tokens, while \(1-M_t^{\mathrm{chunk}}\) indicates the unknown tokens to be reconstructed.

At each reverse denoising step \(k\), we obtain a denoised proposal \(\widetilde{A}^{(k-1)}=\mathcal{D}_{\theta}(A^{(k)},k,C_t)\), noise the reference values to the corresponding level as \(R_t^{(k-1)}=\sqrt{\bar{\alpha}_{k-1}}R_t+\sqrt{1-\bar{\alpha}_{k-1}}\xi_k\), and update the action chunk by replacing the known entries:
\[
A^{(k-1)}
= M_t^{\mathrm{chunk}}\odot R_t^{(k-1)}
+ (1-M_t^{\mathrm{chunk}})\odot\widetilde{A}^{(k-1)},
\]
where \(\mathcal{D}_{\theta}\) denotes one reverse diffusion step using the noise prediction \(f_{\theta}(A^{(k)},k,C_t)\), \(\xi_k\sim\mathcal{N}(0,I)\), and \(C_t\) denotes the conditioning features, including visual features, robot state, and the MAP window.

The final action chunk is obtained by overwriting the known tokens with clean reference values at the last denoising step (\(\bar{\alpha}_0=1\)).
This overwrite operation is applied at every denoising step, ensuring that the final sampled action chunk strictly preserves the specified reference values rather than merely conditioning on them.

\section{EXPERIMENTS}
\label{sec:experiments}
\begin{figure}[t]
\centering
\includegraphics[width=\linewidth]{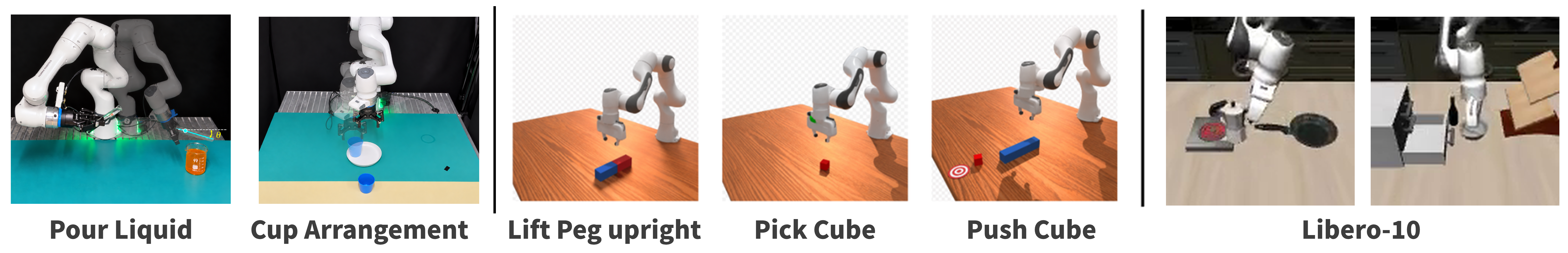}
\caption{Simulation and real-world tasks.}
\label{fig:experiment_tasks}
\end{figure}

Our experiments focus on three questions: (1) What information in physical references matters for action reconstruction?
(2) Can the model generalize to unseen, out-of-distribution mask patterns?
(3) How does each component contribute to the performance of our method?
We evaluate our policy on 15 tasks across ManiSkill, LIBERO-10, and real-world environments, demonstrating diverse manipulation skills and long-horizon capabilities.
Figure~\ref{fig:experiment_tasks} provides an overview of the simulation and real-world tasks.

\paragraph{Dataset Construction.}
For the simulation experiments, we construct action references by masking demonstration trajectories, without collecting additional reference annotations.
Pairing each demonstration with multiple mask configurations provides diverse reference conditions while preserving its observations and target actions.
As shown in Figure~\ref{fig:masked_cartesian_action_space}(C), structured masks reveal modality-specific coordinate groups at selected reference locations for 2D/3D points, 7D poses, and orientations, or retain position coordinates along reference trajectories for 2D/3D flow.
Random masks instead reveal individual scalar action entries across reference locations and action dimensions (Section~\ref{sec:problem_formulation}).
We additionally evaluate \texttt{Mix0}, a composition excluded from joint training that combines dense 2D flow, one 7D pose, and three 3D points (see Supplementary Section~2.4).

\paragraph{Progress Module.}
Our progress module adapts SARM~\citep{chen2026sarmstageawarerewardmodeling} to estimate task progress from frozen CLIP image and text features and proprioceptive states.
The module is trained separately from the policy and uses an eight-layer Transformer with a hidden dimension of 768 and a temporal sampling interval of two frames.

\paragraph{Metrics.}
We report task success rate (SR) to evaluate task completion.
In addition to SR, we introduce \textbf{Control Error (CE)} to quantify how closely the generated trajectory follows the prescribed control signals.
Let $\tau=(a_1,a_2,\ldots,a_T)$ denote the generated action trajectory, where $a_t\in\mathbb{R}^7$, and let $\mathcal{C}=\{c_1,c_2,\ldots,c_K\}$ denote the control signals.
For each signal $c_\ell$, a selection matrix $S_\ell$ extracts the corresponding action dimensions.
We define CE as the average distance from each control signal to its closest matching trajectory point: \( \mathrm{CE}=\frac{1}{K}\sum_{\ell=1}^{K} \min_{t\in\{1,\ldots,T\}} \left\|S_\ell \cdot a_t-c_\ell\right\|_2 \).
Lower CE indicates closer agreement with the prescribed control values.
\paragraph{Baseline.}
We compare our method with the following baselines.
All methods are trained and evaluated using the same data, random seed, and initial states. \textbf{(1) Diffusion Policy (DP).}
We use Diffusion Policy~\citep{chi2025diffusion} with a U-Net backbone as the standard behavior cloning baseline, generating actions without control references.
\textbf{(2) DiT Policy.}
We replace the U-Net with a Diffusion Transformer (DiT)~\citep{peebles2023scalablediffusionmodelstransformers} to isolate the effect of the backbone.
\textbf{(3) Spatiotemporal Conditioning (STC).}
Following prior work on trajectory-, keypose-, and motion-conditioned action generation~\citep{huang2025vlm,nasiriany2025rt,wen2024anypointtrajectorymodelingpolicy,gu2023rt,fan2025diffusion,xu2024flowcrossdomainmanipulationinterface,yu2024bikckeyposeconditionedconsistencypolicy,ma2024hierarchical}, STC flattens spatiotemporal signals and injects them into the diffusion transformer as conditioning features.
It shares the backbone and training hyperparameters of Inpainting Policy to assess the necessity of the Inpainting Policy design.

\subsection{STUDY ON ACTION INPAINTING}
\label{sec:action_inpainting_study}

\begin{figure}[t]
\centering
\includegraphics[width=\linewidth]{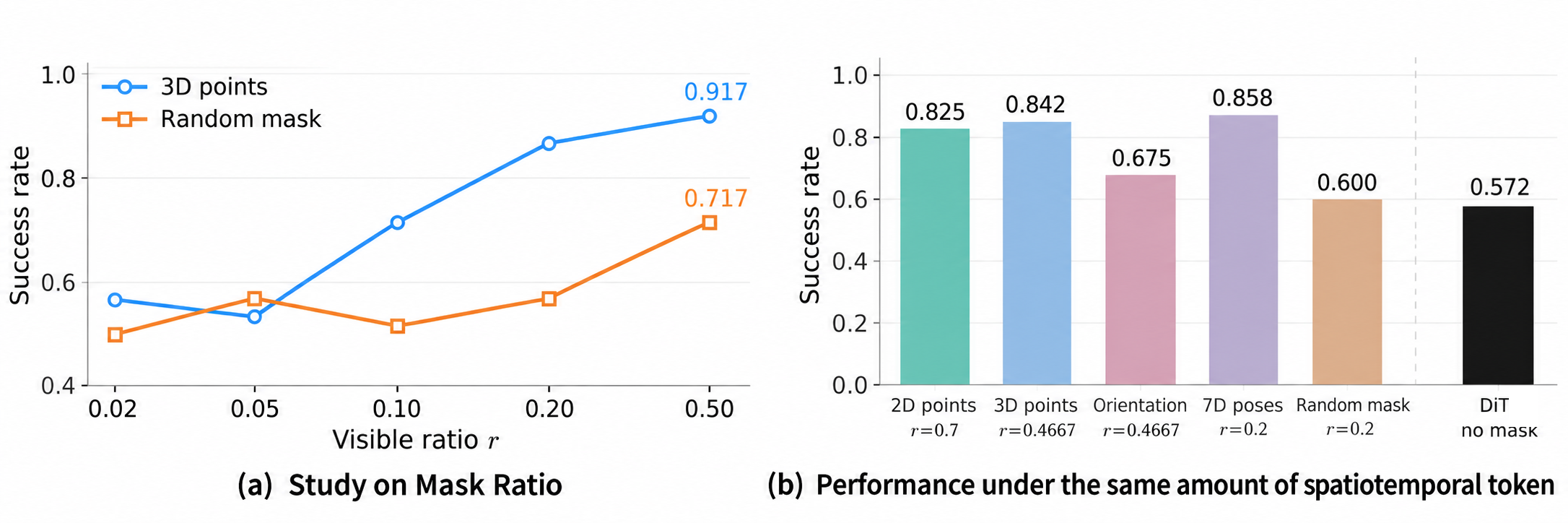}
\caption{Effect of spatiotemporal information quantity and structure.
(a) We vary the visible ratio to examine how the number of references affects task success.
(b) We compare mask types with approximately equal numbers of spatiotemporal tokens, alongside a base policy (a DiT policy), to assess how constraint structure influences the performance of Inpainting Policy.
Here, \(r\) denotes the visible ratio, i.e., the proportion of action information that is not masked in MAP.}
\label{fig:information_quantity_structure}
\end{figure}

As shown in Figure~\ref{fig:information_quantity_structure}, we investigate two aspects of action reconstruction in the temporal Cartesian space: \textbf{(1) Information quantity.}
We examine how the amount of visible control information affects action reconstruction by comparing structured constraints, represented by 3D points, with random constraints.
\textbf{(2) Constraint category.}
We examine the effect of mask structure while approximately matching the number of spatiotemporal tokens across different reference types.
We conduct this study on ManiSkill using 100 demonstrations per task and a DiT backbone with approximately 49M parameters. We vary the visible ratio and mask type and evaluate each configuration on 80 held-out episodes per task. Detailed training settings and mask budgets are provided in Supplementary Sections~2.2 and~2.4, respectively.

\paragraph{Mask Ratio and Mask-Type Analysis.}
\label{sec:inpainting_study_results}
Figure~\ref{fig:information_quantity_structure}(a) shows that increasing the visible-action ratio generally improves task success, indicating that additional known action components provide stronger guidance for reconstructing executable behaviors.
For the structured 3D-point mask, success increases from $57.0\%$ at a ratio of $0.02$ to $91.7\%$ at $0.50$.
Although the random mask also improves with more visible actions, its performance is less stable and reaches only $71.7\%$ at the highest ratio.
This gap suggests that structured control signals convey more task-relevant information than randomly selected action components.

Figure~\ref{fig:information_quantity_structure}(b) compares different mask types under approximately the same number of visible spatiotemporal tokens.
Across all mask types, IP outperforms the unconditioned DiT policy.
Positional and full-pose constraints provide the strongest guidance, whereas orientation-only constraints are less informative for task completion.
The random-mask result further shows that semantically structured action references are more effective than an equivalent number of arbitrary visible tokens.

\subsection{UNIFIED CONTROL VIA JOINT TRAINING}

In this section, as shown in Table~\ref{tab:multitask_single_task_500}, we investigate whether a single policy can follow diverse control references through joint training on versatile spatiotemporal masks.
We assess task completion and reference adherence through success rate and control error.
We evaluate joint training of the Inpainting Policy across diverse constraint types in two settings: single-task policies with a DiT backbone on ManiSkill, and a shared language-conditioned multitask policy with a U-Net backbone on LIBERO-10. IP-RePaint is applied directly at inference time to the baseline DP and DiT policies for Versatile control.

\begin{table}[t]
\caption{\textbf{Unified control via joint training.}
Each policy checkpoint is evaluated across reference types using the average top-three picks of success rate and control error (CE). ManiSkill uses a separate policy for each task, whereas LIBERO-10 uses a shared multitask policy.}
\label{tab:multitask_single_task_500}
\centering
\renewcommand{\arraystretch}{1.10}
\setlength{\tabcolsep}{3.5pt}
\begin{tabular}{l|cc|ccccc}
\toprule
& \multicolumn{2}{c|}{\shortstack{\textbf{Multi-task}\\\textbf{(LIBERO-10)}}}
& \multicolumn{5}{c}{\shortstack{\textbf{Single-task}\\\textbf{(ManiSkill)}}} \\
\cline{2-3}\cline{4-8}
\textbf{Method}
& \makecell[b]{\textbf{Avg. SR}\\\textbf{(\%)} $\uparrow$}
& \makecell[b]{\textbf{CE(all)}$\downarrow$}
& \makecell[b]{\textbf{Lift}\\\textbf{Peg}\\\textbf{SR (\%)}$\uparrow$}
& \makecell[b]{\textbf{Pick}\\\textbf{Cube}\\\textbf{SR (\%)}$\uparrow$}
& \makecell[b]{\textbf{Poke}\\\textbf{Cube}\\\textbf{SR (\%)}$\uparrow$}
& \makecell[b]{\textbf{Avg. SR}\\\textbf{(\%)} $\uparrow$}
& \makecell[b]{\textbf{CE(all)}$\downarrow$} \\
\midrule
DP & 60.00 & --- & 86.33 & 69.33 & 69.67 & 75.11 & --- \\
DiT Policy & 59.67 & --- & 94.67 & 87.33 & \textbf{93.00} & 91.67 & --- \\
STC & 50.67 & 0.060 & 72.67 & 65.67 & 91.67 & 76.67 & 0.041 \\
IP & \textbf{64.67} & 0.045 & \textbf{100.00} & \textbf{90.33} & 87.67 & \textbf{92.67} & 0.023 \\
IP-RePaint & 51.33 & \textbf{0.026} & 95.80 & 82.13 & 62.20 & 80.04 & \textbf{0.014} \\
\bottomrule
\end{tabular}
\end{table}

\paragraph{Joint-Training.}
On LIBERO-10, we adopt \textbf{two-stage joint training}.
First, we train a shared policy across all ten tasks with random masks; second, we continue training with a mixture of structured masks with different reference modalities and sparsity levels. At evaluation time, Mix0 serves as an unseen mask type.
We apply \textbf{mask-wise data augmentation} by pairing each demonstration with multiple mask configurations, yielding diverse reference signals with a limited number of demonstrations.
See Supplementary Sections~2.2 and~2.4 for the training settings and Mix0 construction, and Section~1.4 for the ablation study.

\begin{figure}[t]
\centering
\includegraphics[width=\linewidth]{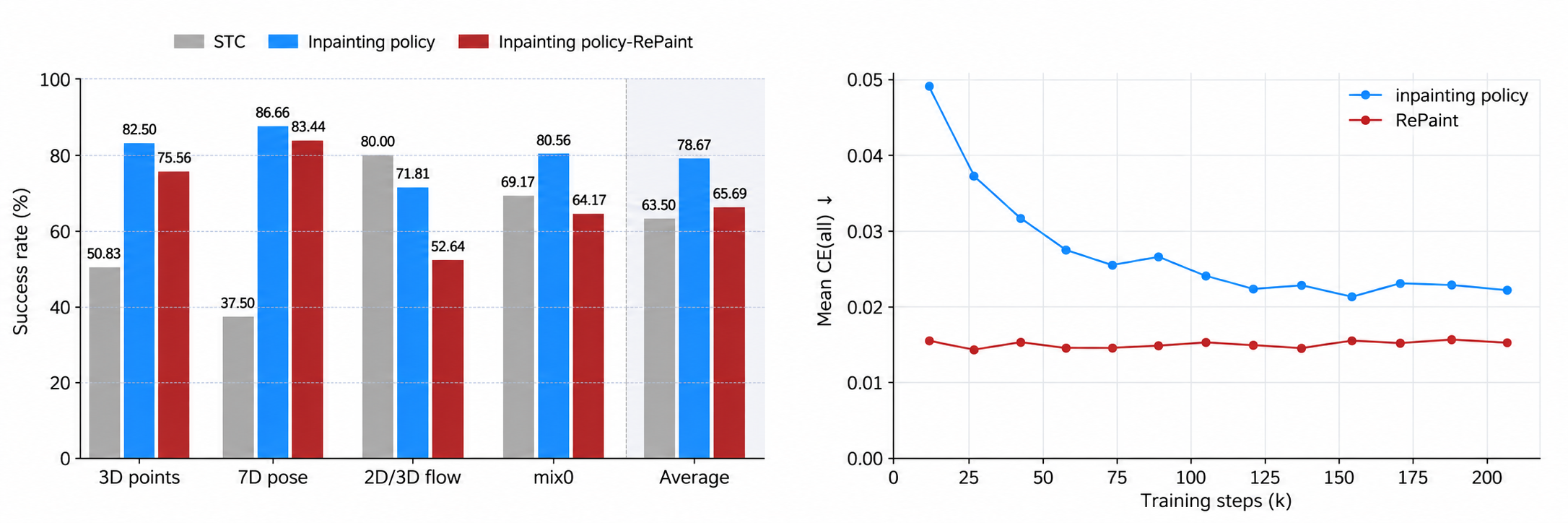}
\caption{\textbf{Mask-type breakdown in joint training.}
Left: We compare success rates across reference types, including the unseen mixed mask \texttt{Mix0}, to evaluate controllability and compositional generalization.
Right: Control error decreases as training progresses.}
\label{fig:mask-type-breakdown}
\end{figure}

\paragraph{Joint-Training Performance.}
Table~\ref{tab:multitask_single_task_500} shows that STC performs worse than its corresponding base policies, indicating that directly conditioning action generation on control signals can interfere with task SR.
In contrast, Inpainting Policy improves task completion while achieving lower control error, performing well in both multi-task and single-task settings and achieving the highest average success rates of $64.67\%$ and $92.67\%$, respectively.
The training-free RePaint variant further reduces CE with relative reductions of $42.2\%$ and $39.1\%$, respectively, demonstrating a trade-off between control accuracy and task completion.

\paragraph{Mask-Type Breakdown and Training Dynamics.}
Figure~\ref{fig:mask-type-breakdown} shows that Inpainting Policy provides the strongest average control performance across all control types, especially on the unseen \texttt{Mix0} mask ($80.56\%$).
STC performs strongly on dense 2D/3D flow ($80.00\%$), but degrades sharply under sparse constraints.
The right plot shows that Inpainting Policy progressively improves its control capability during training, with CE decreasing from approximately $0.049$ to $0.022$.
In contrast, RePaint maintains the lowest CE, around $0.014$--$0.015$, across all checkpoints.
Notably, since the \texttt{Mix0} composition is absent from the training set, its high success rate indicates that the policy can \textbf{compose learned mask patterns} without additional finetuning.

\subsection{CONTROL WITH REAL-WORLD REFERENCES}

\paragraph{Real World Setting.}
For real-world experiments, we use Franka R3 robotic arms with two-finger grippers.
One calibrated RealSense D455 camera provides the third-person view, and one ZED-M camera provides the wrist view.
The robotic arm operates at a policy frequency of 15 Hz using absolute end-effector commands.
We train task-specific Inpainting Policies (IP) and Diffusion Policy (DP) as the base policy.
See Supplementary Section~2.6 for real-world implementation details.

\begin{figure}[t]
\centering
\includegraphics[width=\linewidth]{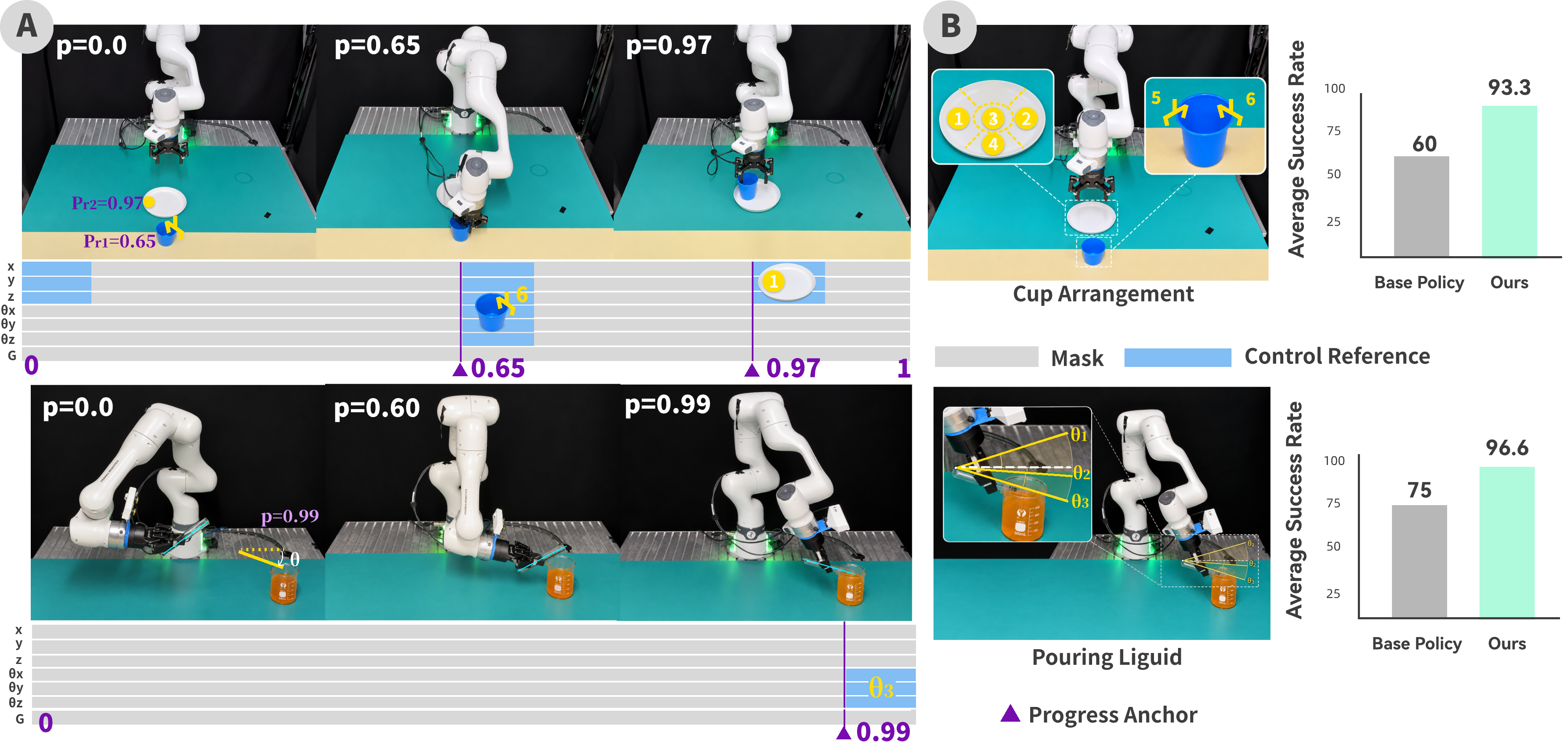}
\caption{\textbf{Real-world spatiotemporal editing.}
Here, \(p_r\) denotes the progress associated with a user-provided reference, whereas \(p\) denotes the robot's current execution progress.
The rollouts and corresponding MAP masks illustrate how these references guide execution.}
\label{fig:real_world_control}
\end{figure}

\paragraph{Human-in-the-Loop Control.}
The \textbf{Robotic Spatiotemporal Editing} system accepts a variable number of user-provided references, each associated with a reference type and a progress value $p'$ that specifies when the constraint should be applied.
The system converts all references into a unified \textbf{Masked Action Space (MAP)} for policy inference.

\paragraph{Tasks.}
We evaluate whether IP can complete the tasks while successfully satisfying additional spatial and orientational objectives with references including keypoints, flows, poses, and orientations, under both sparse and dense configurations.
As shown in Figure~\ref{fig:real_world_control} (B), in the cup-arrangement task, these references determine how the cup is grasped and where it is placed on the plate.
In the liquid-pouring task, orientation references specify the tube's tilt angle and thereby regulate the dispensed volume.

\paragraph{Result.}
As shown in Figure~\ref{fig:real_world_control}, with three or more reference points, the policy achieves success rates above \(93\%\), while the cup-to-reference-point distance ranges from \(18.2\) to \(63.7\,\mathrm{mm}\), with an average of \(44.3\,\mathrm{mm}\).
In the liquid-pouring task, the tube-tilt error remains below \(8^\circ\), with a mean of \(7.8^\circ\).
These results show that providing explicit control constraints substantially improves task completion and reference adherence: denser references produce smoother motion and smaller deviations from the target.
In contrast, when only a single reference is provided at the final progress point, the task completion rate falls below $30\%$, revealing a limitation of sparse late-stage supervision and an important direction for improving the subsequent editing system.
Supplementary Section~2.6 provides additional recorded SR and normalized CE results (Tables~8--10), plus deployment tips.

\section{CONCLUSION AND FUTURE WORK}
\label{sec:conclusion}
\paragraph{Conclusion.}
We presented a unified framework for spatiotemporal action editing that augments policies with DiT or U-Net backbones with a unified control interface over heterogeneous spatiotemporal references.
Without requiring additional control-specific data or increasing the backbone training budget, our method preserves the original task completion ability while enabling users to edit and compose spatiotemporal action constraints through masked action inpainting.
Our inpainting formulation is model-agnostic and may be compatible with a broad range of foundation action models, including WAM and flow-matching-based VLAs, which we leave for future empirical study.

\paragraph{Limitations.}
Our current inpainting implementation may not be optimal; alternatives such as CFG and ControlNet-style conditioning remain unexplored.
Future work will study large-scale BERT-style random-mask pretraining for foundation action models and evaluate zero-shot editing under more unstructured control specifications.

\clearpage
\section*{AI USE DISCLOSURE}

In this work, we used generative AI tools to assist with translating and polishing the manuscript, drafting and editing parts of the paper, improving the readability of the presentation, modifying scientific figures, interpreting experimental results, and writing and modifying code.

We did not use generative AI tools to generate synthetic datasets, develop theoretical models or conceptual frameworks, formulate mathematical claims, provide critical ingredients for proving mathematical claims, write proofs, propose or refine hypotheses, design or provide feedback on the research methodology or experiments, clean or reformat datasets, or perform qualitative or thematic data analysis.

All AI-assisted content was reviewed and verified by the authors.
The authors take full responsibility for the final content of this submission, including all text, claims, experimental results, code, and figures.

\section*{ETHICS STATEMENT}

This work studies controllable robot manipulation using public simulation benchmarks and controlled real-world robotic experiments.
It does not involve human subjects, personal data, sensitive information, or high-risk applications.
The real-world experiments were conducted in a controlled laboratory environment following standard robot-safety procedures.
We have read and followed the ICLR Code of Ethics and take responsibility for the ethical conduct of this research.

\section*{REPRODUCIBILITY STATEMENT}

To support reproducibility, the supplementary material describes the masked-action formulation and inference procedure in \textit{Additional Method Details}, and the policy architectures, training settings, mask construction, evaluation metrics, and real-world setup in \textit{Additional Implementation Details}.
The \textit{Ablation Studies} section reports the settings and evaluation protocols for the encoder, window-length, progress-module, training-schedule, and progress-alignment comparisons.
These descriptions distinguish comparisons with matched settings from historical comparisons with differing training or evaluation conditions, and identify results aggregated across reference configurations and checkpoints.

\clearpage

\bibliography{references}
\bibliographystyle{iclr2027_conference}

\clearpage
\suppressfloats[t]
\setcounter{section}{0}
\renewcommand{\theHsection}{supp.\arabic{section}}

\begin{center}
    {\Large\bfseries Supplementary Material}
\end{center}

\section{Ablation Studies}
\label{app:experiments}

We study how the MAP encoder, forward window length, progress module, training schedule, and progress source affect Inpainting Policy (IP).
These comparisons inform the design choices in the main paper.
Some settings differ in more than one factor, so their results cannot isolate each factor's contribution; we describe these differences alongside each experiment.
Top-$k$ SR is the average of the $k$ highest evaluated SR values within the same run.
CE(all), CE(success), and CE(failed) denote control error over all, successful, and failed rollouts, respectively; lower CE is better.

\subsection{MAP Window Encoding and Feature Dimension}
\label{app:window_encoding}

On ManiSkill PickCube, we compare a 1DConv encoder with a 32-dimensional output against 2DConv encoders with output dimensions of 32, 64, and 128.
Each output forms the MAP conditioning vector.
All runs use 100 demonstrations, seed 1, observation/execution/prediction horizons of 2/8/16, a forward window of 32, and the \texttt{points@1} mask configuration.
The IP Transformer uses a hidden dimension of 512, six encoder and six decoder blocks, and a feed-forward dimension of 2048.

\begin{table}[htbp]
\caption{\textbf{MAP encoder and feature-dimension comparison on PickCube.}
Training budgets, batch sizes, and evaluation counts are not fully matched.
Bold marks the highest recorded Top-3 SR.}
\label{tab:app_encoding}
\centering
\begin{tabular}{lccc}
\hline
\noalign{\vskip 3pt}
\textbf{Encoder} & \shortstack{\textbf{Feature}\\\textbf{dimension}} & \shortstack{\textbf{Top-3 SR}\\(\%) $\uparrow$} & \shortstack{\textbf{Last evaluation}\\\textbf{step}} \\
\hline
\noalign{\vskip 3pt}
1DConv & 32 & 12.00 & 155k \\
2DConv & 32 & 19.00 & 180k \\
2DConv & 64 & 14.67 & 175k \\
2DConv & 128 & \textbf{22.00} & 275k \\
\hline
\end{tabular}
\end{table}

At output dimension 32, 2DConv reaches a Top-3 SR of 19.00\%, compared with 12.00\% for 1DConv (Table~\ref{tab:app_encoding}).
The 2DConv results are 19.00\%, 14.67\%, and 22.00\% at dimensions 32, 64, and 128, respectively.
The 128-dimensional encoder used in the main paper has the highest recorded SR, but it is also evaluated later, at 275k updates.
The results therefore reflect differences in both the encoder and its training configuration.

\subsection{MAP Window Length}
\label{app:window_length}

We test forward MAP windows of 52, 64, and 70 reference locations on ManiSkill PickCube-v1, keeping the training and evaluation settings fixed.
Each model uses the same 100 demonstrations, 200k training updates, a batch size of 128, observation/execution/prediction horizons of 2/8/16, and mean-squared-error loss with mean reduction.
The backward window length is zero, and the forward window includes the current reference anchor; its length is separate from the IP prediction horizon.
All three models use a 128-dimensional 2DConv encoder, the \texttt{points@1} mask configuration, and the same STPM checkpoint, \texttt{reward\_best.pt}.

\begin{table}[htbp]
\caption{\textbf{MAP window-length comparison on PickCube-v1.}
SR statistics summarize three training seeds; end SR measures final-step success.
CE(all) is the aggregate error at 200k.
Bold marks the highest mean success rates and lowest CE.}
\label{tab:app_window_length}
\centering
\begin{tabular}{ccccc}
\hline
\noalign{\vskip 3pt}
\shortstack{\textbf{Forward}\\\textbf{length}} & \shortstack{\textbf{Top-3 SR}\\(\%) $\uparrow$} & \shortstack{\textbf{200k SR}\\(\%) $\uparrow$} & \shortstack{\textbf{200k end SR}\\(\%) $\uparrow$} & \shortstack{\textbf{200k}\\\textbf{CE(all)} $\downarrow$} \\
\hline
\noalign{\vskip 3pt}
52 & \textbf{70.22 $\boldsymbol{\pm}$ 4.25} & \textbf{63.33 $\boldsymbol{\pm}$ 3.79} & \textbf{36.33 $\boldsymbol{\pm}$ 3.06} & \textbf{0.02317} \\
64 & 61.33 $\pm$ 3.00 & 55.67 $\pm$ 5.03 & 33.33 $\pm$ 4.93 & 0.03281 \\
70 & 60.22 $\pm$ 3.40 & 55.33 $\pm$ 2.89 & 32.33 $\pm$ 4.73 & 0.03198 \\
\hline
\end{tabular}
\end{table}

The 52-location window performs best across the reported metrics (Table~\ref{tab:app_window_length}).
Its Top-3 SR is 70.22\%, exceeding the 64- and 70-location windows by 8.89 and 10.00 percentage points, respectively.
At the final checkpoint, it also has the highest SR (63.33\%) and end SR (36.33\%), along with the lowest CE(all) (0.02317).
Both the Top-3 and final-checkpoint results favor a forward length of 52 among the settings tested on PickCube-v1.
Even with the same mask rule, however, window length can affect both temporal coverage and how much reference information is visible.

\subsection{Progress-Module Configuration}
\label{app:stpm_configuration}

To compare three STPM configurations on LIBERO-10, we keep the IP checkpoint fixed at 190k and replace only STPM, without retraining IP.
We evaluate each configuration on 50 episodes per seed, using evaluation seeds 1000, 2000, and 3000.
Table~\ref{tab:app_stpm} lists each module's architecture, observation history, frame gap, and checkpoint-selection criterion.

\begin{table}[htbp]
\caption{\textbf{STPM configuration comparison on LIBERO-10 with a fixed 190k IP checkpoint.}
SR is mean $\pm$ population standard deviation over three evaluation seeds.
Architecture, observation history, and checkpoint selection vary together.
Bold marks the highest mean SR.}
\label{tab:app_stpm}
\centering
\begin{tabular}{ccccc}
\hline
\noalign{\vskip 3pt}
\textbf{STPM} & \shortstack{\textbf{Dimension /}\\\textbf{layers / heads}} & \shortstack{\textbf{Obs. /}\\\textbf{gap}} & \shortstack{\textbf{Checkpoint}\\\textbf{selection}} & \textbf{SR (\%)} $\uparrow$ \\
\hline
\noalign{\vskip 3pt}
v3 & 512 / 4 / 8 & 1 / 1 & Sequence best & \textbf{74.00 $\boldsymbol{\pm}$ 0.00} \\
v5 & 768 / 8 / 12 & 6 / 2 & Endpoint best & 70.67 $\pm$ 5.73 \\
v6 & 544 / 5 / 8 & 6 / 2 & Endpoint best & 69.33 $\pm$ 3.40 \\
\hline
\end{tabular}
\end{table}

STPM v3 gives the highest mean SR at 74.00\%, followed by v5 at 70.67\% and v6 at 69.33\%.
All three modules support task completion, with mean SRs between 69.33\% and 74.00\%.
The larger models offer no improvement for this IP checkpoint, although the comparison also changes observation history and checkpoint selection along with architecture.

\subsection{Joint and Two-Stage Training}
\label{app:training_schedule}

We compare joint and two-stage training on LIBERO-10 using 500 source demonstrations and 50 evaluation trajectories.
Both runs use STPM v4, a forward window of 64, a 128-dimensional MAP encoding, a U-Net IP backbone, and seed 1000, with the short-window option disabled.
Joint training uses a four-slot reference mixture throughout steps 0--180k.
Two-stage training uses random masks for steps 0--90k, then switches to the same four-slot mixture for steps 90k--180k.
We also report an independent two-stage run with 1,000 source demonstrations (100 per task), 50 evaluation episodes, a global batch size of 96, and STPM v4 in Table~\ref{tab:app_training_schedule}.
This run uses the same short-window setting, forward window length, and MAP feature dimension.

\begin{table}[htbp]
\caption{\textbf{Training-schedule comparison on LIBERO-10.}
All SR values are percentages; Top-1 step identifies the best evaluated checkpoint.
Bold marks the higher SR within the 500-demonstration comparison.
The 1,000-demonstration run is an independent experiment.}
\label{tab:app_training_schedule}
\centering
\begin{tabular}{clcccc}
\hline
\noalign{\vskip 3pt}
\shortstack{\textbf{Source}\\\textbf{demos}} & \textbf{Training} & \shortstack{\textbf{Top-1 SR}\\$\uparrow$} & \shortstack{\textbf{Top-1}\\\textbf{step}} & \shortstack{\textbf{Top-3 SR}\\$\uparrow$} & \shortstack{\textbf{Top-5 SR}\\$\uparrow$} \\
\hline
\noalign{\vskip 3pt}
500 & Joint & 56.00 & 180k & 50.00 & 47.60 \\
500 & Two-stage & \textbf{62.00} & 105k & \textbf{56.00} & \textbf{54.40} \\
\hline
\noalign{\vskip 3pt}
1,000 & Two-stage & 100.00 & 150k & 97.33 & 96.00 \\
\hline
\end{tabular}
\end{table}

With 500 demonstrations, two-stage training reaches Top-3 and Top-5 SRs of 56.00\% and 54.40\%, compared with 50.00\% and 47.60\% for joint training.
The improvement supports the schedule used in the main paper: random-mask pretraining followed by structured joint training.
The separate run with 1,000 demonstrations reaches its best evaluated SR of 100.00\% at 150k, with Top-3 and Top-5 SRs of 97.33\% and 96.00\%.
This result suggests a benefit from more demonstrations, though dataset size is not the only factor that differs between the independent runs.

\subsection{Estimated Progress versus Rollout Timestep}
\label{app:progress_source}

We compare STPM-based alignment with a rollout-timestep baseline on PickCube, using the same IP checkpoint, 100 evaluation trajectories, seed 1, and a maximum episode length of 160.
For learned alignment, we use the STPM checkpoint \texttt{STPM\_pickcube/reward\_best.pt}.
The baseline selects the MAP window using $p_t'=t/(160-1)$, where $t$ is the zero-based rollout step, in place of the estimated progress.
Both modes retain the progress column in the reference trajectory.

\begin{table}[htbp]
\caption{\textbf{Progress-source comparison with a fixed PickCube IP checkpoint and 100 evaluation trajectories.}
SR is reported as a percentage.
Bold marks the higher success rate or lower CE in each column.
Successful and failed subsets can differ between modes.}
\label{tab:app_progress_source}
\centering
\begin{tabular}{lcccc}
\hline
\noalign{\vskip 3pt}
\shortstack[l]{\textbf{Progress}\\\textbf{source}} & \shortstack{\textbf{SR}\\$\uparrow$} &  \shortstack{\textbf{CE(all)}\\$\downarrow$} & \shortstack{\textbf{CE(success)}\\$\downarrow$} & \shortstack{\textbf{CE(failed)}\\$\downarrow$} \\
\hline
\noalign{\vskip 3pt}
STPM & \textbf{83.00} & \textbf{0.02912} & \textbf{0.02675} & \textbf{0.04073} \\
Rollout timestep & 67.00 & 0.03983 & 0.03297 & 0.05377 \\
\hline
\end{tabular}
\end{table}

STPM alignment raises success-once SR from 67\% to 83\%, a gain of 16 percentage points, while reducing CE(all) from 0.03983 to 0.02912 (Table~\ref{tab:app_progress_source}).
CE is also lower for both successful and failed rollouts, consistent with closer reference adherence, though the rollouts in each subset can differ between modes.
For this checkpoint and evaluation set, learned progress alignment improves both task success and overall reference accuracy.

\section{Additional Implementation Details}
\label{app:experimental_details}

\subsection{Benchmarks, Data, and Baselines}
\label{app:benchmarks_baselines}

\paragraph{Benchmarks and dataset construction.}
ManiSkill covers picking and placing objects, controlling object rotation, and pushing objects to target locations.
LIBERO-10 comprises ten long-horizon manipulation tasks with visual observations and language instructions.
Figure 3 (Simulation and real-world tasks) in the main paper summarizes the simulation and real-world tasks.
We construct training and test examples by masking demonstrations, without collecting additional control references.
Mask-wise data augmentation pairs each demonstration with multiple mask configurations while preserving its observations and target actions.

\paragraph{DP and DiT Policy.}
DP uses a U-Net backbone for observation-to-action generation without an explicit spatiotemporal control interface~\citep{chi2025diffusion}.
DiT Policy replaces the U-Net with a Diffusion Transformer~\citep{peebles2023scalablediffusionmodelstransformers}, keeping the action prediction objective and training protocol unchanged to examine the effect of the backbone.

\paragraph{Spatiotemporal Conditioning (STC).}
Following methods that condition action generation on trajectories, keyposes, or motion guidance~\citep{huang2025vlm,nasiriany2025rt,wen2024anypointtrajectorymodelingpolicy,gu2023rt,fan2025diffusion,xu2024flowcrossdomainmanipulationinterface,yu2024bikckeyposeconditionedconsistencypolicy,ma2024hierarchical}, STC flattens spatiotemporal control signals and injects them into a Diffusion Transformer as conditioning features.
It shares the backbone and training hyperparameters of Inpainting Policy, allowing a comparison between direct reference conditioning and masked-action reconstruction.
Within each comparison, methods use the same data, random seed, and initial states.
We additionally evaluate IP-RePaint as a training-free inference variant of Inpainting Policy.

\subsection{Policy Architectures and Training}
\label{app:training_details}
\paragraph{Mask ratio.}
\label{sec:mask_ratio}
We investigate how the amount of visible control information affects action reconstruction by comparing structured constraints, represented by 3D points, with random constraints.
We quantify the information provided by either constraint type using the spatiotemporal information budget $N_{\mathrm{ST}}$, the number of visible scalar action entries.
For a reference sequence of length $T$ with action dimension $D=7$, 3D-point and random constraints provide approximately $3rT$ and $rTD$ visible entries, respectively.
Here, the structured-mask ratio is defined over reference locations, whereas the random-mask ratio is defined over scalar action entries, following the mask settings shown in the main paper.
Equal visible ratios therefore do not imply equal information budgets; Section~\ref{app:mask_construction} gives the precise definitions and budget-matched settings.

\paragraph{Mask category.}
\label{sec:mask_category}
We next examine the effect of constraint structure while approximately matching the number of spatiotemporal tokens across constraint types.
Each token corresponds to one visible action coordinate at one reference location.
Thus, a 2D point contains two tokens, a 3D point contains three, and a 7D pose contains seven; complete 2D and 3D trajectories contain $2T$ and $3T$ tokens, respectively.
Random constraints contribute one token per revealed action coordinate.
We set $N_{\mathrm{ST}}=0.2TD=1.4T$ and use visible ratios of $0.70$, $7/15$, $0.20$, and $0.20$ for 2D points, 3D points, 7D poses, and random masks, respectively.
An orientation reference specifies three rotational coordinates and therefore uses the same visible ratio as a 3D point, $r_{\mathrm{orientation}}=7/15$.
For random constraints, the ratio is instead computed over all action entries: $r_{\mathrm{random}}=N_{\mathrm{ST}}/(TD)$.
\paragraph{Action-inpainting study.}
The mask study uses a DiT with six encoder blocks, six decoder blocks, a hidden dimension of 512, and approximately 49M trainable parameters.
Each policy is trained on 100 demonstrations per task for 100,000 iterations, with a batch size of 64 and a learning rate of $10^{-4}$.
Each visibility and mask-type configuration is evaluated on 80 held-out episodes per task.

\paragraph{Single-task joint training on ManiSkill.}
Each task uses a separate policy trained on 500 demonstrations with heterogeneous spatiotemporal masks.
The DiT architecture is the same as in the mask study.
Training uses 300,000 updates, a global batch size of 128, and a learning rate of $5\times10^{-5}$.

\paragraph{Multi-task joint training on LIBERO-10.}
We train a shared language-conditioned policy on the official LIBERO-10 dataset.
The policy uses a U-Net-based Diffusion Policy with channel dimensions of 512, 1024, and 2048, and a lightweight Transformer language encoder.
Two-stage joint training first uses random masks across all ten tasks and then continues with a mixture of structured masks spanning different guidance modalities and sparsity levels.
Training uses 180,000 updates, a global batch size of 96, and a learning rate of $10^{-4}$.
The fixed test set contains 100 trajectories derived from the official LIBERO-10 pruned states; masking these trajectories produces the evaluation references.

\paragraph{Progress module.}
We adapt SARM~\citep{chen2026sarmstageawarerewardmodeling} to estimate task progress from frozen CLIP image and text features and proprioceptive states.
The module is trained separately and uses an eight-layer Transformer with a hidden dimension of 768 and a temporal sampling interval of two frames.
We refer to the progress module as STPM in the configuration descriptions and comparisons below.

\subsection{Control-Error Definitions}
\label{app:control_error}

\paragraph{Simulation.}
Let $\tau=(a_1,\ldots,a_T)$ be a generated action trajectory, with $a_t\in\mathbb{R}^{7}$, and let $\mathcal{C}=\{c_1,\ldots,c_K\}$ contain the prescribed control signals.
A selection matrix $S_\ell$ extracts the action dimensions specified by $c_\ell$.
We define
\begin{equation}
    \mathrm{CE}
    =
    \frac{1}{K}
    \sum_{\ell=1}^{K}
    \min_{t\in\{1,\ldots,T\}}
    \left\|S_\ell a_t-c_\ell\right\|_2.
    \label{eq:control_error}
\end{equation}
Lower CE indicates closer agreement with the reference values.
Because each signal is matched to its closest trajectory point, this metric measures reference proximity rather than alignment to a prescribed timestep.
CE is not reported for the unconditioned DP and DiT baselines.

\paragraph{Real-world evaluation.}
For the real-world results in Section 4.3 of the main paper, we report task success rate, cup-to-reference-point distance in millimeters, and tube-tilt error in degrees.
The distance and angle errors quantify task-specific physical outcomes; they are not interchangeable with normalized action-space CE.
We report normalized CE separately from these physical errors.
The evaluated tasks, reference configurations, and reported result ranges are described in Section~\ref{app:real_world_details}.

\subsection{Mask Construction and Information Budgets}
\label{app:mask_construction}

\paragraph{Information budget.}
For a binary visibility mask $\mathbf{M}\in\{0,1\}^{T\times D}$, where $T$ denotes the reference sequence length, $D=7$, and $M_{p,d}=1$ indicates a visible coordinate, the spatiotemporal information budget is
\begin{equation}
    N_{\mathrm{ST}}
    =
    \|\mathbf{M}\|_0
    =
    \sum_{p=0}^{T-1}
    \sum_{d=1}^{D}
    M_{p,d}.
\end{equation}
A spatiotemporal token denotes one visible action coordinate at one reference location.
Thus, 2D points, 3D points, and 7D poses contain two, three, and seven tokens per selected reference location.
Complete 2D and 3D trajectories contain $2T$ and $3T$ tokens, respectively.
Random masks reveal individual action entries.

\paragraph{Visibility-ratio study.}
For 3D-point constraints, a visible ratio $r$ reveals approximately $3rT$ scalar entries.
For random masks, the same ratio reveals approximately $rTD$ entries.
Equal visible ratios therefore do not imply equal information budgets.

\paragraph{Budget-matched mask comparison.}
We set $N_{\mathrm{ST}}=0.2TD=1.4T$.
For structured references revealing $d$ coordinates per selected reference location, the corresponding visible ratio is
\begin{equation}
    r_d
    =
    \frac{N_{\mathrm{ST}}}{dT}.
\end{equation}
For random constraints, the ratio is computed over all action entries: $r_{\mathrm{random}}=N_{\mathrm{ST}}/(TD)$.
Table~\ref{tab:mask_budgets} lists the resulting ratios.
Budgets are matched approximately because the numbers of selected reference locations and entries are discrete.

\begin{table}[t]
    \caption{ \textbf{Visible ratios for the budget-matched mask comparison, with $N_{\mathrm{ST}}=1.4T$.}
    Structured-mask ratios are defined over reference locations; the random-mask ratio is defined over scalar action entries.
    }
    \label{tab:mask_budgets}
    \centering
    \begin{tabular}{lcc}
        \textbf{Reference type}
        & \shortstack{\textbf{Coordinates per}\\\textbf{reference location}}
        & \textbf{Visible ratio} \\
        \hline
        \noalign{\vskip 3pt}
        2D point
        & 2 & $0.70$ \\

        3D point
        & 3 & $7/15$ \\

        Orientation
        & 3 & $7/15$ \\

        7D pose
        & 7 & $0.20$ \\

        Random
        & --- & $0.20$ \\
    \end{tabular}
\end{table}

\paragraph{Unseen mixed mask: \texttt{Mix0}.}
This pattern is excluded from joint training.
It preserves $(x,y)$ over the entire trajectory.
At four randomly selected distinct reference locations, it additionally retains one full 7-dimensional action at one location and an $(x,y,z)$ position at each of the other three locations.
The resulting mask combines dense 2D flow, one 7D pose, and three 3D points, as described in the main paper.
Evaluation uses the trained policy without fine-tuning.

\subsection{Additional Joint-Training Results}
\label{app:additional_joint_results}

\paragraph{Per-task performance.}
Table~\ref{tab:maniskill_per_task} reports per-task results on ManiSkill: Inpainting Policy achieves the highest SR on \textsc{LiftPegUpright} and \textsc{PickCube}, while DiT Policy performs best on \textsc{PokeCube}.

\begin{table}[t]
    \caption{ \textbf{Per-task ManiSkill success rates (\%).}
    Each task uses a separate policy trained with 500 demonstrations.
    }
    \label{tab:maniskill_per_task}
    \centering
    \begin{tabular}{lccc}
        \textbf{Method}
        & \shortstack{\textbf{LiftPeg}\\\textbf{Upright}}
        & \shortstack{\textbf{Pick}\\\textbf{Cube}}
        & \shortstack{\textbf{Poke}\\\textbf{Cube}} \\
        \hline
        \noalign{\vskip 3pt}
        DP
        & 86.33 & 69.33 & 69.67 \\

        DiT Policy
        & 94.67 & 87.33 & \textbf{93.00} \\

        STC
        & 72.67 & 65.67 & 91.67 \\

        IP
        & \textbf{100.00} & \textbf{90.33} & 87.67 \\

        IP-RePaint
        & 95.80 & 82.13 & 62.20 \\
    \end{tabular}
\end{table}

\paragraph{Mask-type breakdown.}
The evaluation in Figure 5 (Mask-type breakdown in joint training) in the main paper reports an average SR of $78.67\%$ for Inpainting Policy and an SR of $80.56\%$ on the unseen \texttt{Mix0} mask.
IP-RePaint achieves $52.64\%$ on dense 2D/3D flow and $64.17\%$ on \texttt{Mix0}, illustrating the trade-off between reference adherence and task completion described in the main paper.

\paragraph{Training dynamics.}
In Figure 5 (Mask-type breakdown in joint training) in the main paper, Inpainting Policy's CE decreases from approximately $0.049$ to $0.022$ during training.
IP-RePaint maintains the lowest CE, approximately $0.014$--$0.015$, across the evaluated checkpoints.
In the aggregate results in Table 1 (Unified control via joint training) in the main paper, IP-RePaint reduces CE relative to Inpainting Policy by $42.2\%$ on LIBERO-10 and $39.1\%$ on ManiSkill, with lower SR in both settings.

\subsection{Real-World Setup and Evaluation}
\label{app:real_world_details}

\paragraph{Hardware and observations.}
We use Franka R3 robotic arms equipped with two-finger grippers.
A calibrated RealSense D455 camera provides the third-person view, and a ZED-M camera provides the wrist view.
The policy operates at 15\,Hz and outputs absolute end-effector commands.

\paragraph{Policies and training.}
We train task-specific Diffusion Policy (DP) baselines and Inpainting Policies (IP) on the same demonstration datasets.
Both use frozen, pretrained DINOv2 ViT-S/14 encoders for dual-camera RGB observations, together with proprioceptive inputs and a U-Net denoiser with channel dimensions of 256, 512, and 1024.
At inference time, IP additionally receives the user-provided action references through a 32-step MAP window.
DP serves as the base policy for observation-to-action generation.

\paragraph{Human-in-the-loop control.}
The Robotic Spatiotemporal Editing system accepts a variable number of user-provided references.
Each reference specifies a reference type, the corresponding action values, and a normalized progress value $p'\in[0,1]$ indicating the task stage at which the constraint should apply.
The system converts these inputs into visibility masks and partially specified action tokens in the Masked Action Space (MAP) interface.
At inference time, a progress model estimates the current task-progress anchor and selects the corresponding MAP window.
IP conditions on this window, visual observations, and proprioceptive state to reconstruct the unspecified action dimensions and predict a complete action chunk for execution.
Here, MAP denotes the reference interface, and IP denotes the learned policy used by the interactive system.

\paragraph{Tasks and reference configurations.}
We evaluate whether references allow IP to satisfy additional spatial and orientational objectives while completing the underlying manipulation task.
Across the real-world evaluation, the system uses heterogeneous reference types, including keypoints, flow, poses, and orientations, under both sparse and dense reference configurations.
In cup arrangement, the underlying objective is to pick up a cup and place it on a plate; the references additionally control how the cup is grasped and where it is placed.
In controlled test-tube pouring, the underlying objective is to move a test tube above a beaker and pour liquid; orientation references specify the tube's tilt angle to regulate the dispensed volume.
Figure 6 in the main paper illustrates the reference-guided behaviors and summarizes the reported outcomes.

\paragraph{Evaluation metrics.}
We report task success rate together with task-specific physical reference errors.
For cup arrangement, the physical error is the cup-to-reference-point distance in millimeters; for pouring, it is the tube-tilt error in degrees.
These quantities describe task outcomes and are distinct from normalized action-space CE, as discussed in Section~\ref{app:control_error}.
The rollouts also provide qualitative evidence about motion smoothness and response to the requested references.
The per-configuration tables below additionally report normalized action-space CE from the recorded evaluations.
These CE values are dimensionless and should not be interpreted as millimeter distances or tilt angles.

\paragraph{Cup arrangement results.}
For the real-world results described in the main paper, IP achieves success rates above $93\%$ with three or more reference points, and the cup-to-reference-point distance ranges from $18.2$ to $63.7\,\mathrm{mm}$, with a mean of $44.3\,\mathrm{mm}$.
With only a single reference at the final progress point, task success falls below $30\%$.
Table~\ref{tab:supp_real_cup} reports success rates, normalized CE, and final TCP error across cup-reference configurations.
These additional evaluations span different reference configurations and checkpoints; their scope differs from the results summarized in the main paper.
IP reaches 100\% SR with four 3D reference points at 20k updates and eight 2D reference points at 15k updates.
For the 15k 3D configurations, increasing $K$ from 2 to 4 is accompanied by an SR increase from 12.50\% to 87.50\% and a CE decrease from 0.175191 to 0.063748.
The mean final TCP error is $32.60\,\mathrm{mm}$ for the 20k 3D configuration with $K=4$, and $41.29$ and $36.53\,\mathrm{mm}$ for the 15k 3D configurations with $K=2$ and $K=4$, respectively.
For the 15k 2D configurations, the corresponding errors are $66.29$, $56.35$, and $59.16\,\mathrm{mm}$ for $K=4$, $8$, and $16$, respectively (Table~\ref{tab:supp_real_cup}).
The results support using references across relevant task stages, while success also depends on reference type, placement, and policy checkpoint.
The exact SR and CE for the single-point configuration are not available in this table.

\begin{table}[htbp]
\caption{\textbf{Cup arrangement results by reference configuration.}
IP denotes Inpainting Policy without RePaint; $K$ is the number of reference points.
Checkpoints are identified by their training step, with k denoting 1,000 updates; a dash indicates an unreported or inapplicable value.
Positional error is the 3D distance from the last recorded TCP position to the reference endpoint, averaged equally over all rollouts in each configuration, including failures, and indicating the positional error of the cup.}
\label{tab:supp_real_cup}
\centering
\begin{tabular}{llccccc}
\textbf{Policy} & \textbf{Reference} & $K$
& \textbf{Checkpoint}
& \shortstack{\textbf{SR}\\$(\%)\uparrow$}
& \shortstack{\textbf{Norm.}\\\textbf{CE}$\downarrow$}
& \shortstack{\textbf{Positional}\\\textbf{error (mm)}$\downarrow$} \\
\hline
\noalign{\vskip 3pt}
IP & 3D position & 4 & \shortstack{20k} & \textbf{100.00} & 0.078249 & 32.60 \\
IP & 2D position & 16 & \shortstack{15k} & 80.00 & 0.069302 & 59.16 \\
IP & \shortstack[l]{3D position} & 4 & \shortstack{15k} & 87.50 & 0.063748 & 36.53 \\
IP & \shortstack[l]{3D position} & 2 & \shortstack{15k} & 12.50 & 0.175191 & 41.29 \\
IP & 2D position & 4 & \shortstack{15k} & 0.00 & 0.125435 & 66.29 \\
IP & 2D position & 8 & \shortstack{15k} & \textbf{100.00} & 0.048606 & 56.35 \\
IP & 3D position & 1 & \shortstack{15k} & --- & --- & --- \\
DP & None & --- & \shortstack{100k} & 50.00 & --- & --- \\
\end{tabular}
\end{table}

\paragraph{Liquid-pouring results.}
Table~\ref{tab:supp_real_water} shows that IP achieves 93.33\% SR with four orientation references and 100\% with eight or sixteen references.
The reported angle errors range from $7.6867^{\circ}$ to $7.8133^{\circ}$, consistent with the below-$8^{\circ}$ error and mean error of $7.8^{\circ}$ reported in the main paper.
Normalized CE varies non-monotonically with reference count, with the lowest value of 0.090101 at $K=16$.
Thus, task completion, normalized reference error, and physical tilt error capture different aspects of control performance.

\begin{table}[htbp]
\caption{\textbf{Controlled test-tube pouring results by reference count.}
All IP rows use the \texttt{orientation0.01 / STPM} configuration without RePaint; $K$ is the number of orientation references.
DP receives no reference.
The DP success labels remain unverified; angle errors are means for the corresponding reference configurations.}
\label{tab:supp_real_water}
\centering
\begin{tabular}{lccccc}
\textbf{Policy} & $K$ & \textbf{Checkpoint}
& \shortstack{\textbf{SR}\\$(\%)\uparrow$}
& \shortstack{\textbf{Norm.}\\\textbf{CE}$\downarrow$}
& \shortstack{\textbf{Angle error}\\$(^{\circ})\downarrow$} \\
\hline
\noalign{\vskip 3pt}
IP & 4 & \shortstack{30k} & 93.33 & 0.094315 & 7.8133 \\
IP & 8 & \shortstack{30k} & \textbf{100.00} & 0.101605 & 7.6867 \\
IP & 16 & \shortstack{30k} & \textbf{100.00} & 0.090101 & 7.7567 \\
DP & --- & \shortstack{50k} & 43.48 & --- & --- \\
\end{tabular}
\end{table}

\paragraph{Aggregate results.}
Table~\ref{tab:supp_real_aggregate} summarizes the recorded configurations, yielding IP success rates of 77.19\% for cup arrangement and 97.78\% for pouring, with normalized CE values of 0.086105 and 0.095340, respectively.
The corresponding DP success rates are 50\% and 43.48\%, with the latter subject to the label-verification limitation noted in Table~\ref{tab:supp_real_water}.
The IP aggregates pool the additional reference configurations and checkpoints reported here; they do not represent the performance of one fixed policy configuration.
The high-success results described in the main paper refer to its evaluated reference-guided settings, while these tables provide additional results across configurations.

\begin{table}[htbp]
\caption{\textbf{Aggregate real-world results across recorded reference configurations.}
Cup denotes cup arrangement; Water denotes controlled test-tube pouring.
IP excludes RePaint.
The Water DP success labels remain unverified.}
\label{tab:supp_real_aggregate}
\centering
\begin{tabular}{llcc}
\textbf{Task} & \textbf{Policy}
& \shortstack{\textbf{SR}\\$(\%)\uparrow$}
& \shortstack{\textbf{Norm.}\\\textbf{CE}$\downarrow$}\\
\hline
\noalign{\vskip 3pt}
Cup & IP & \textbf{77.19} & 0.086105 \\
Cup & DP & 50.00 & --- \\
Water & IP & \textbf{97.78} & 0.095340 \\
Water & DP & 43.48 & --- \\
\end{tabular}
\end{table}

\paragraph{Practical tips for real-robot deployment.}
Progress used for window selection should evolve smoothly and monotonically.
Discontinuous or backward updates can skip reference points or reactivate constraints from an earlier task stage.
A calibrated mapping from predicted progress to reference position, with limits on backward motion and abrupt advances, helps maintain consistent window selection.
Progress completion should be distinguished from successful physical task completion.

References should cover the task stages that matter for the requested behavior.
For cup arrangement, guidance during grasping and transport can complement a final placement target; for pouring, orientation references can specify the intended progression of tube tilt.
Because IP receives a local MAP window, a single late-stage reference may not become visible until execution approaches its associated progress value.

Reference positions and orientations should remain compatible with the training distribution and with one another.
Requests far outside the demonstrated range can fail even when they are representable through MAP.
IP reconstructs actions using learned behavior, so encoding a constraint in the interface does not by itself guarantee reliable execution of every requested behavior.

\section{Additional Method Details}
\label{app:method_details}

Throughout this section, $t$ indexes policy execution, $p$ indexes reference locations, and $k$ indexes diffusion steps.

\subsection{Progress-Guided MAP Alignment}
\label{app:map_alignment_details}

\paragraph{Progress estimation.}
We use normalized task progress rather than the raw execution timestep to align control references with policy execution.
Given the current observation $o_t=(I_t,s_t)$, where $I_t$ denotes camera observations and $s_t$ denotes the robot state, a separately trained progress estimator predicts
\[
p_t' = g_\phi(o_t), \qquad p_t'\in[0,1].
\]
Using estimated progress to select references accounts for variations in execution speed and aligns references with the corresponding task stages.

\paragraph{Mapping progress to the reference trajectory.}
Let $A^{\mathrm{ref}}\in\mathbb{R}^{T_{\mathrm{ref}}\times7}$ denote the reference action-value array at progress locations $p_p'$, where $p=0,\ldots,T_{\mathrm{ref}}-1$.
Each row is an action vector $a_p=(x_p,y_p,z_p,\theta_{x,p},\theta_{y,p},\theta_{z,p},g_p)$, while each scalar entry is a spatiotemporal action token.
The binary visibility mask $M\in\{0,1\}^{T_{\mathrm{ref}}\times7}$ identifies specified entries, and $V=M\odot A^{\mathrm{ref}}$ contains the masked reference values.
The current progress estimate selects the nearest reference index $p_t=\arg\min_p |p_p'-p_t'|$.
For uniformly spaced reference progress values $p_p'=p/(T_{\mathrm{ref}}-1)$ with $T_{\mathrm{ref}}>1$, this becomes
\begin{equation}
    p_t
    =
    \tau(p_t')
    =
    \left\lfloor
    p_t' (T_{\mathrm{ref}}-1)
    \right\rceil.
\end{equation}
Given $h$ preceding reference locations and $j$ additional future reference locations, we select
\begin{equation}
\mathcal{W}_t
=
\mathrm{MAP\_Window}(p_t)
=
\left[
a_{p_t-h},
\ldots,
a_{p_t},
\ldots,
a_{p_t+j}
\right],
\label{eq:map_window}
\end{equation}
with indices clipped to $[0,T_{\mathrm{ref}}-1]$.
For out-of-range locations, we repeat the nearest boundary action and its visibility mask, preserving a fixed window length of $h+j+1$.
The forward portion contains $j+1$ locations, including the current anchor.
Thus, a reported forward window length of $L_{\mathrm{forward}}$ corresponds to $j=L_{\mathrm{forward}}-1$ and is distinct from the prediction horizon $H$.

This formulation avoids directly using normalized progress $p_t'$ as a discrete trajectory index and explicitly defines the correspondence between task progress and the reference sequence.

\paragraph{MAP encoding.}
Although $\mathcal{W}_t$ is written as an action sequence for clarity, the encoder receives only the corresponding masked values and visibility masks.
Let $V_t^{\mathrm{win}}\in\mathbb{R}^{(h+j+1)\times7}$ and $M_t^{\mathrm{win}}\in\{0,1\}^{(h+j+1)\times7}$ denote the rows selected from $V$ and $M$ using the same indices and boundary handling as $\mathcal{W}_t$.
We provide these representations to a lightweight 2D convolutional encoder, keeping the visibility mask explicit so that an unspecified entry can be distinguished from a specified zero value.
The encoder produces a 128-dimensional MAP feature vector, which is concatenated with visual and proprioceptive features to condition the denoising network.

The local window serves two purposes.
First, it associates spatial constraints with their corresponding interaction stage.
Second, it supplies nearby reference context whose temporal coverage is controlled by $h$ and $j$.
Section~\ref{app:window_length} evaluates the effect of forward window length, and Section~\ref{app:progress_source} compares estimated progress with rollout-timestep alignment.

\subsection{RePaint Inference}
\label{app:repaint_inference}

The standard Inpainting Policy (IP) generates a complete action chunk conditioned on observations and the local MAP window.
We additionally consider a RePaint-style inference strategy~\citep{lugmayr2022repaintinpaintingusingdenoising} that reinserts specified reference values during denoising without additional policy training.

Let $R_t\in\mathbb{R}^{H\times7}$ contain the clean reference values aligned with the target action chunk, and let $M_t^{\mathrm{chunk}}\in\{0,1\}^{H\times7}$ identify the specified entries.
Only entries selected by $M_t^{\mathrm{chunk}}$ are used for replacement; the remaining reference values are not required.
The reference values and sampled actions use the same action normalization.
Sampling starts from Gaussian noise $A^{(K)}\sim\mathcal{N}(0,I)$, where $K$ is the number of reverse diffusion steps.
At each step $k$, we first obtain a denoised proposal $\widetilde{A}^{(k-1)}=\mathcal{D}_{\theta}(A^{(k)},k,C_t)$ and noise the reference values to the corresponding level as $R_t^{(k-1)}=\sqrt{\bar{\alpha}_{k-1}}R_t+\sqrt{1-\bar{\alpha}_{k-1}}\xi_k$.
Here, $\mathcal{D}_{\theta}$ denotes one reverse diffusion step using the noise prediction $\epsilon_{\theta}(A^{(k)},k,C_t)$, $\xi_k\sim\mathcal{N}(0,I)$, and $C_t$ contains the observation and local MAP conditioning features.
The reverse diffusion step produces an action proposal at noise level $k-1$; the noise prediction itself is not an action sample.

We retain the proposal at unspecified entries and replace the specified entries with the noised reference values:
\begin{equation}
A^{(k-1)}
=
M_t^{\mathrm{chunk}}\odot R_t^{(k-1)}
+
(1-M_t^{\mathrm{chunk}})\odot\widetilde{A}^{(k-1)}.
\label{eq:repaint_action}
\end{equation}

This replacement is applied at every reverse diffusion step.
At the last step, we use clean reference values with $\bar{\alpha}_0=1$, so the final sampled action chunk satisfies $M_t^{\mathrm{chunk}}\odot A^{(0)}=M_t^{\mathrm{chunk}}\odot R_t$.
This equality concerns the sampled action values and does not guarantee exact reference satisfaction during physical execution.

The learned Inpainting Policy and its RePaint inference variant are evaluated separately in our experiments: the former learns to reconstruct actions through the conditional denoising objective, whereas the latter adds explicit reference replacement during sampling.

\subsection{Masked Denoising Objective}
\label{app:masked_denoising}

Given a clean demonstration action chunk $A^{(0)}\in\mathbb{R}^{H\times7}$, a diffusion step $k$, and Gaussian noise $\epsilon\sim\mathcal{N}(0,I)$, we construct
\[
A^{(k)}
=
\sqrt{\bar{\alpha}_k}A^{(0)}
+
\sqrt{1-\bar{\alpha}_k}\epsilon.
\]
Here, $\bar{\alpha}_k$ is the cumulative signal-retention coefficient.

The denoising network predicts $\widehat{\epsilon}=\epsilon_{\theta}(A^{(k)},k,C_t)$ from the noisy action chunk and the conditioning features defined in Section~\ref{app:conditioning_architecture}.
The visibility mask $M_t^{\mathrm{chunk}}\in\{0,1\}^{H\times7}$ is aligned with the target action chunk.
It is distinct from the full reference mask $M$ and the local conditioning-window mask $M_t^{\mathrm{win}}$, whose temporal lengths are $T_{\mathrm{ref}}$ and $h+j+1$, respectively.
We optimize
\begin{equation}
\mathcal{L}_{\mathrm{MAP}}
=
\mathbb{E}
\left[
\begin{aligned}
& w_m \left\|M_t^{\mathrm{chunk}}\odot(\epsilon-\widehat{\epsilon})\right\|_F^2
\\
& {}+ w_u \left\|(1-M_t^{\mathrm{chunk}})\odot(\epsilon-\widehat{\epsilon})\right\|_F^2
\end{aligned}
\right],
\label{eq:map_loss_appendix}
\end{equation}
where $w_m$ and $w_u$ weight the visible and missing regions, and $\|\cdot\|_F^2$ sums squared errors over the action-chunk entries.
The expectation is over training examples, sampled masks, diffusion steps, and Gaussian noise.

The unknown-token objective trains the model to infer actions that are not specified by the reference.
The visible-token objective encourages consistency between predicted actions and specified control values.
This differs from RePaint inference, where visible values are explicitly overwritten rather than enforced only through the learned denoising model.

\subsection{Mask Families for Joint Training}
\label{app:mask_training}

We construct multiple masked-action examples from each demonstration.
The training distribution contains three complementary mask families: structured masks ($\alpha$), no-reference masks ($\beta$), and random masks ($\gamma$).

\paragraph{$\alpha$: Structured control masks.}
The structured family represents control references that naturally arise from external planning or human specification.
It includes:
\begin{itemize}
    \item 2D position or trajectory constraints;
    \item 3D position or trajectory constraints;
    \item 7D poses comprising Cartesian position, orientation, and gripper state;
    \item orientation-only constraints;
    \item gripper-state constraints; and
    \item sparse keypoints or keyposes.
\end{itemize}

These masks expose semantically meaningful subsets of the Cartesian action trajectory and teach the policy to interpret different reference modalities within a shared representation.
The first three coordinates specify Cartesian position, the next three specify orientation, and the seventh specifies the gripper state.
Here, 2D references constrain Cartesian $x,y$ coordinates, while flow references specify position trajectories across reference locations.
Image-space references must first be mapped into the action coordinate system.
For a reference at progress $p'$, we select $p^\star=\arg\min_p|p_p'-p'|$ and reveal the corresponding action coordinates in row $p^\star$ of $M$.
For example, a 3D point reveals coordinates $\{1,2,3\}$, while an orientation reference reveals $\{4,5,6\}$.

\paragraph{$\beta$: No-reference masks.}
For this family, the visibility ratio is zero:
\[
M=\mathbf{0}.
\]
The policy therefore generates actions without an external action reference while retaining its observation conditioning.
Including this case trains the policy to operate when action references are unavailable.

\paragraph{$\gamma$: Random token masks.}
Random masks reveal randomly selected scalar action entries across reference locations and action dimensions.
Unlike the structured family, these patterns are not tied to a predefined control modality.
They expose the model to a substantially larger space of partial action specifications and encourage token-level reconstruction beyond the structured patterns seen during training.

We sample a mask family and then a mask conditioned on that family:
\[
c\sim\rho(c), \quad c\in\{\alpha,\beta,\gamma\}, \qquad
M\sim p(M\mid c),
\]
and apply $M$ to the demonstration action trajectory to construct the reference condition for diffusion training.
The distribution $\rho(c)$ controls the relative frequency of structured, no-reference, and random-mask examples.
For LIBERO-10, the two-stage schedule first uses random masks and then continues with a mixture of structured masks spanning different reference modalities and sparsity levels.
The training settings and schedule comparison are described in Sections~\ref{app:training_details} and~\ref{app:training_schedule}, respectively.

\subsection{Generalization to Unseen Mask Compositions}
\label{app:mask_generalization}

Because all control signals are represented as visible entries in the same $T_{\mathrm{ref}}\times7$ reference action space, compatible reference modalities can be combined by taking the union of their visible tokens.
For two masks $M_1$ and $M_2$ defined on the same reference progress grid, their composition is
\begin{equation}
M_{\mathrm{mix}}
=
M_1 \lor M_2,
\end{equation}
with the corresponding visible action entries assembled from the two references.
This composition assumes that any overlapping specified entries agree.

This representation does not introduce a modality-specific input head or control interface.
Consequently, the policy can accept a mask composition that does not appear explicitly in the training set.
Our experiments evaluate its ability to follow such compositions using \texttt{Mix0}, a mixed-mask pattern excluded from training, without additional policy fine-tuning.
The construction of \texttt{Mix0} is defined in Section~\ref{app:mask_construction}.

\subsection{Conditioning Architecture}
\label{app:conditioning_architecture}

The denoising model is conditioned on three information sources:
\begin{equation}
C_t
=
\left[
\phi_{\mathrm{vis}}(I_t),
\phi_{\mathrm{state}}(s_t),
\phi_{\mathrm{MAP}}
\left(
M_t^{\mathrm{win}},
V_t^{\mathrm{win}}
\right)
\right].
\end{equation}

Here, $\phi_{\mathrm{vis}}$ extracts visual features, $\phi_{\mathrm{state}}$ encodes robot proprioception, and $\phi_{\mathrm{MAP}}$ embeds the local masked-action window defined in Section~\ref{app:map_alignment_details}.
The MAP encoder uses 2D convolutions followed by pooling and projection to produce a 128-dimensional feature vector.
The brackets denote concatenation of this vector with the visual and proprioceptive features.
The resulting condition $C_t$ is supplied to $\epsilon_{\theta}(A^{(k)},k,C_t)$ during action denoising.
The encoder comparison is reported in Section~\ref{app:window_encoding}.

The formulation is agnostic to the specific diffusion backbone.
In our experiments, we instantiate it using either a Diffusion Transformer or a U-Net-based Diffusion Policy depending on the benchmark.
See Section~\ref{app:training_details} for policy architectures and training settings.

\end{document}